\documentclass[11pt,a4paper]{article}
\usepackage[hyperref]{emnlp2020}

\usepackage[utf8]{inputenc} % allow utf-8 input
\usepackage{microtype}
\usepackage{graphicx}
\usepackage{subfigure}
\usepackage{booktabs} 
\usepackage{comment}
\usepackage{hyperref}
\usepackage{mathtools}
\usepackage{color}
\usepackage{graphicx}
\usepackage{pifont}
\usepackage{CJKutf8}
\usepackage{url}
\usepackage{multicol,multirow}
\usepackage{makecell}
\usepackage{array}
\usepackage{bm}
\usepackage{parskip}
\usepackage[T1]{fontenc}    % use 8-bit T1 fonts
\usepackage{booktabs}       % professional-quality tables
\usepackage{amsmath, amsthm}
\usepackage{amssymb}
\usepackage{amsfonts}

\usepackage{times}
\usepackage{float}
\usepackage{stfloats}
\usepackage{latexsym}

\usepackage{microtype}
\definecolor{ao}{rgb}{0.0, 0.5, 0.0}
\definecolor{asparagus}{rgb}{0.53, 0.66, 0.42}
\definecolor{amber}{rgb}{1.0, 0.49, 0.0}
\definecolor{alizarin}{rgb}{0.82, 0.1, 0.26}
\definecolor{applegreen}{rgb}{0.55, 0.71, 0.0}
\definecolor{amethyst}{rgb}{0.6, 0.4, 0.8}
\definecolor{auburn}{rgb}{0.43, 0.21, 0.1}
\aclfinalcopy % Uncomment this line for the final submission
\title{OpenViDial 2.0: A Larger-Scale, Open-Domain Dialogue Generation\\Dataset with Visual Contexts}
\date{}

\author{
Shuhe Wang$^{\clubsuit\spadesuit}$, 
Yuxian Meng$^\clubsuit$, 
Xiaoya Li$^\clubsuit$\\
{\bf Xiaofei Sun$^\clubsuit$,
Rongbin Ouyang$^\spadesuit$
Jiwei Li$^{\blacklozenge\clubsuit}$}\\
$^\clubsuit$ Shannon.AI, $^\spadesuit$Peking University, $^\blacklozenge$Zhejiang University\\ 
  \{shuhe\_wang, yuxian\_meng, xiaoya\_li, xiaofei\_sun, jiwei\_li\}@shannonai.com\\

}

\begin{document}
\maketitle

\begin{abstract}
In order to better simulate the real human conversation process, models need to generate dialogue utterances based on not only preceding textual contexts but also visual contexts. However, with the development of multi-modal dialogue learning, the dataset scale gradually becomes a bottleneck.
In this report, we release {\bf OpenViDial 2.0}, a larger-scale open-domain multi-modal dialogue dataset compared to the previous version OpenViDial 1.0 \citep{meng2020openvidial}. OpenViDial 2.0 contains a total number of 5.6 million dialogue turns extracted from either movies or TV series from different resources, and each dialogue turn is paired with its corresponding visual context.
We hope this large-scale dataset can help facilitate future researches on open-domain multi-modal dialog generation, e.g., multi-modal pretraining for dialogue generation.\footnote{Dataset is available found at 
{\url{https://github.com/ShannonAI/OpenViDial}}.}

\end{abstract}

\section{Introduction}

Developing open-domain dialogue agents is of growing interest \citep{,li2017adversarial,ghazvininejad2017knowledge,zhou2017emotional,gao2018neural,asghar2018affective,han2020non,zhou2020design}. Existing methods for developing effective open-domain dialogue agents mostly follow a two-step pipeline: (1) collecting a large-scale dataset containing massive dialog turns from real conversations, and (2) training a neural model to learn to generate high quality responses given the previous dialogue contexts \cite{li2016deep,li2016persona,zhang2018personalizing,huang2020challenges}.

Since most methods are data-driven, a large-scale and high quality open-domain dialogue datasets may be the first matter to be considered before designing the model. \citet{meng2020openvidial} released the OpenViDial dataset which contains a total number of 1.1 million dialogue turns with utterances paired with visual context. Some recent works leveraged the OpenViDial dataset and built effective multi-modal dialog models \citep{wang2021modeling} on top, demonstrating that learning multi-modal features gives rise to higher response quality.

In this report, we collect and extend OpenViDial, releasing OpenViDial 2.0, a much larger-scale open-domain dialogue dataset with visual contexts. In common with the prior version OpenViDial 1.0 \citep{meng2020openvidial}, the dialogue turns and visual contexts in OpenViDial 2.0 are also extracted from movies and TV series, where each dialogue turn is paired with the corresponding visual context in which it takes place. OpenViDial 2.0 contains a total number of 5.6 million dialogue turns along with 5.6 million visual contexts stored as images, a scale of 4 times larger than OpenViDial 1.0. 
We hope this large-scale dataset can help facilitate future researches on open-domain multi-modal dialog generation, e.g., multi-modal pretraining for dialogue generation. 

\section{Related Work}
\subsection{Open Domain Dialog Datasets}

\paragraph{Textual Dialog Datasets} Since the task of open-domain dialog generation has developed for many years, there are various open-domain dialog datasets only consists textual information. For simulating the movie conversation, there are OpenSubtitle dataset \citep{Tiedemann2009NewsFO, Lison2016OpenSubtitles2016EL} and Cornell Movie-Dialogs Corpus \citep{Danescu-Niculescu-Mizil+Lee:11a}. The OpenSubtitle dataset is a large-scale dataset contains a total number of 3.35G sentence fragments extracted from the OpenSubtitle website, while the Cornell Movie-Dialogs Corpus contains a collection of movie conversations extracted from raw movie scripts. For simulating the social conversation, there are PersonaChat \cite{zhang2018personalizing} and Twitter Triple Corpus \cite{sordoni2015neural}. The Twitter Triple Corpus consists of 4,232 Twitter conversation triples evaluated from 33K candidate triples by human raters. Other datasets such as the Ubuntu Dialog Corpus \cite{lowe2015ubuntu} and EmpatheticDialogues \cite{rashkin2018towards} are both commonly used for textual open-domain dialog generation.

\paragraph{Visual Dialog Datasets} A mount of datasets containing visual features have been developed, since the task of VisualDialog is first introduced by \citet{das2017visual}, where a model is required to answer questions by given a dialog history and the image itself as contexts. For this work, \citet{das2017visual} released VisDial v0.9 and v1.0 datasets which contains 120K images from MSCOCO\footnote{\url{http://mscoco.org/}} and each image is associated with 10 rounds of question-answer dialog. Further, other datasets like the GuessWhat?! dataset \cite{devries2017guesswhat}, the CLEVERDialog dataset \cite{kottur2019clevrdialog}, the MNIST-Dialog dataset \cite{seo2017visual} and the Audio Visual Scene-Aware Dialog (AVSD) dataset \citep{hori2018endtoend,alamri2019audio} are mainly focus more on answering questions according to an image or video rather than dialogue generation with visual contexts.The OpenViDial dataset \cite{meng2020openvidial} is released to alleviate this situation, where contains 1.1M dialogue turns and each dialogue turn paired with the corresponding visual context in which it takes place. And thus, models need to learn to generate dialogue utterances not only based on preceding textual contexts but also visual contexts.

\subsection{Dialog Generation}

\paragraph{Open Domain Dialog Generation} Open-domain dialog generation is a simulation for real human conversations and is a traditional task in NLP \citep{weizenbaum1966eliza, COLBY197537, Wallace2009}. Currently, the most researches for open-domain dialog generation are based on sequence-to-sequence architecture \citep{vinyals2015neural, li2015diversity, dodge2016evaluating, serban2016hierarchical, zhao2017learning, xie2017data, lee2019convlab, ghandeharioun2019approximating, li2020teaching, han2020explaining, zhang2019dialogpt, roller2020recipes}. And whether a model can generate diverse \citep{xu2018dp, baheti2018generating}, coherent \citep{li2016deep,li2017adversarial,tian-etal-2017-make,bosselut-etal-2018-discourse,adiwardana2020towards}, informative \citep{shao2017generating,lewis-etal-2017-deal,ghazvininejad2017knowledge,young2017augmenting,zhao2019rethinking} and knowledge-fused \citep{10.1145/3340531.3411967,zhao-etal-2020-knowledge-grounded,he-etal-2020-amalgamating} responses or not has become metrics to evaluate a dialog generation model. However, the mainly researches described above are developed on textual only and the development of multi-modal dialog generation is relatively slow since the lack of large-scale datasets.

\paragraph{Visual Dialog Generation} Most of existing works apply attention mechanisms to model the interplay between text and visual contexts \citep{lu2017best, kottur2018visual, jiang-bansal-2019-self, yang2019making, guo2019dual, niu2019recursive, kang2019dual, park2020multi, jiang2020dam}. Other techniques like reinforcement learning \citep{das2017learning, wu2018you}, variational auto-encoders \cite{massiceti2018flipdial} and graph networks \citep{zheng2019reasoning, jiang2020kbgn} have also been employed to the visual dialog task. More recently, based on the OpenViDial dataset \cite{meng2020openvidial}, \citet{wang2021modeling} proposed three attention-based models \cite{vaswani2017attention} to generate dialogue utterances given the preceding text-visual contexts and further proposed to build text-visual dependency to improve the dialogue quality, making an initial step for the task of text-visual open-domain dialogue generation rather than answering questions based on an image.

\begin{table*}
\centering
\begin{tabular}{lcc} \\\toprule
\textbf{Statistics} & \textbf{OpenViDial 1.0} & \textbf{OpenViDial 2.0} \\\midrule
Number of turns & 1.1M & 5.6M \\
Number of images & 1.1M & 5.6M \\
Vocab size before BPE & 70K & 278K \\
Vocab size after BPE & 30K & 30K \\
Average length of each episode & 14 & 48 \\
Average length of each turn & 7.6 & 8.3 \\\bottomrule
\end{tabular}
\caption{Detailed statistics for OpenViDial 2.0 and a comparison to OpenViDial 1.0.}
\label{tab:stats}
\end{table*}

\begin{table}[t]
\centering
\begin{tabular}{lcc} \\\toprule
& \textbf{OpenViDial 1.0} & \textbf{OpenViDial 2.0} \\\midrule
Train & 1M & 4.6M \\
Dev & 50K & 0.5M \\
Test & 50K & 0.5M \\\bottomrule
\end{tabular}
\caption{Splitting for training, dev and test}
\label{tab:split}
\end{table}

\section{Constructing OpenViDial 2.0}
In this section, we describe the details of constructing of OpenViDial 2.0. We first collect a raw dataset consisting of about 800 English movies and TV series with an average length of 2.5 hours per video. Each video has a corresponding external English subtitle file where each line is a string including the subtitle text and the time interval. There is no video embedded with any internal subtitles.

The full process to construct OpenViDial 2.0 can be divided into three steps: (1) segmenting each video into multiple frames; (2) pairing each frame with subtitle text from it corresponding subtitle file; (3) splitting these (image, text) pairs into different dialog turns. The OpenCV \cite{opencv_library} toolkit is used to segment each video into multiple images by frame, and we discard the initial and the last 10 minutes of each video because of the general existence of intro in movies and TV series. To pair images with textual subtitles for each video, we first read the video's subtitle file row-by-row and obtain the time interval as well as the subtitle text. Then, we extract a group of images according to the time interval, and randomly choose one image from the group as the visual context paired with the subtitle text, forming a paired (image, text) dialog turn.

We are able to construct a final dataset of 5.6M dialog turns, where each turn consists of a sequence of words and an image. The size of the image is one of (1) 1280$\times$720, (2) 1920$\times$1080, and (3) 2048$\times$1080 according to different video resources. We employ the BPE tokenizer \cite{sennrich-etal-2016-neural} to preprocess the text. A detailed comparison with OpenViDial 1.0 is shown in Table \ref{tab:stats}. The splitting for training, dev and test is shown in Table \ref{tab:split}.

\begin{table*}
\centering
\small
\scalebox{0.85}{
\begin{tabular}{lcccc} \\\toprule
{\bf Dataset} & {\bf Genre} & {\bf Multi-Modal?} & {\bf \# Sentences} &  {\bf \# Images}\\\midrule 
OpenSubtitles 2016 \citep{Lison2016OpenSubtitles2016EL} & Plain-text Dialog & \ding{55} &337M & -- \\
Cornell Movie-Dialogs \citep{Danescu-Niculescu-Mizil+Lee:11a} & Plain-text Dialog & \ding{55}  &0.3M &--  \\
VisDial v1.0 \citep{das2017visual}  & VQA & \checkmark & 2.4M & 120K  \\
Guess-What?! \citep{devries2017guesswhat}  & VQA & \checkmark & 0.8M & 66K \\
AVSD \citep{alamri2019audio} &  VQA & \checkmark & 152K & -- \\
OpenViDial 1.0 \citep{meng2020openvidial} & Visual+Text Dialog &  \checkmark & 1.1M & 1.1M  \\
OpenViDial 2.0 & Visual+Text Dialog &  \checkmark & 5.6M & 5.6M  \\\bottomrule
\end{tabular}
}
\caption{A comparison of different datasets. VQA: Visual Question Answering.}
\label{tab:comparison}
\end{table*}

In Table \ref{tab:comparison}, we make a comparison with existing widely-used dialog datasets. Both OpenViDial 1.0 and OpenViDial 2.0 focus on multi-modal dialog generation in comparison to VisDial, Guess-What?! and AVSD which focus more on VQA.
Comparing against OpenViDial 1.0, OpenViDial 2.0 is much larger in scale,  about 5 times as big as OpenViDial 1.0.

\begin{table*}
  \small
  \centering
  \begin{tabular}{l|cccccc} \toprule
  \multicolumn{1}{l}{\bf System} & \multicolumn{1}{l}{\bf Model} & \multicolumn{1}{l}{\bf BLEU} & 
  \multicolumn{1}{l}{\bf Dis-1} & \multicolumn{1}{l}{\bf Dis-2} & \multicolumn{1}{l}{\bf Dis-3} & \multicolumn{1}{l}{\bf Dis-4}\\\midrule 
  \multirow{2}{*}{\bf NV} & {\it w/o MI} & \text{1.95} & \text{0.0037} & \text{0.0302} & \text{0.0929}  & \text{0.1711}\\
   & {\it w/ MI} & \text{1.96} & \text{0.0039} & \text{0.0311} & \text{0.0953}  & \text{0.1630}\\\hline
   \multirow{2}{*}{CV} & {\it w/o MI} & \text{1.97} & \text{0.0041} & \text{0.0353} & \text{0.0999} & \text{0.1726}\\
    & {\it w/ MI} & \text{1.98} & \text{0.0047} & \text{0.0392} & \text{0.1093} & \text{0.1774} \\\hline
   \multirow{2}{*}{FV} & {\it w/o MI} & \text{1.99} & \text{0.0056} & \text{0.0431} & \text{0.1250} & \text{0.2215}\\
    & {\it w/ MI} & \text{2.00} & \text{0.0060} & \text{0.0460} & \text{0.1321} & \text{0.2311}\\
  \bottomrule
  \end{tabular}
  \caption{Automatic evaluation results for BLEU, Stopword\% and Diversity. }
  \label{tab:bleu}
\end{table*}

To evaluate OpenViDial 2.0, we experiment on OpenViDial2.0 using multi-modal dialog models proposed in \cite{wang2021modeling}. 

\subsection{Vanilla Visual Dialog Models}
According to the granularity of the visual features ranges from none, coarse-grained image features to fine-grained object features, \citet{wang2021modeling} proposed three vanilla visual dialog models: (1) the NoVisual(NV) model, (2) the CoarseVisual(CV) model and (3) the FineVisual(FV) model.

\paragraph{NoVisual} The NV model is a general uni-modal dialog generation model, which is required to learn to generate responses using only dialog texts without visual information. A standard Transformer \cite{vaswani2017attention} architecture is used as the backbone for the NV model. For each dialog turn, all the preceding dialog texts are packed into a long sequence with a special token as the delimiter. Then, this sequence is embedded with positional encodings including sentence-level positional encoding and token-level positional encoding. Last, it is fed to the Transfromer as input.

\paragraph{CoarseVisual} In contrast to the NV model, the CV model injects coarse-level visual information into dialog generation. For each dialog turn, it utilizes a ResNet-50 model \cite{kaiming2016resnet} pre-trained on ImageNet \cite{krizhevsky2012imagenet} to extract a high-dimensional feature for each image as the visual information. Then the image feature is added to its corresponding text representation forming the text-visual feature. Positional encodings are also used to notify position information. The concatenated long text-visual sequence is fed into the Transformer model.

\paragraph{FineVisual} Extracting visual information from a coarse view might be insufficient to model fine-grained visual elements in images such as facial expressions, body gestures as well as physical motions. The FV model thus uses Faster R-CNN \cite{ren2015faster} pre-trained on Visual Genome \cite{krishna2017visual} to extract fine-grained visual features. Different from the CV model, the FV model directly concatenates the set of extracted fine-grained visual information with the dialog texts into a long sequence. And except for  the sentence-level and token-level positional embeddings, there is an additional positional embedding for visual features. 

\subsection{Visual-Text Mutual Dependeny}
Although each response is generated according to the preceding textual and visual contexts, there is no guarantee on whether or how much the visual contexts are used. To significantly strength the connection between the generated response and its visual contexts, \citet{wang2021modeling} proposed to model the mutual information (MI) between visual contexts and text features. To put it simply, we use visual feature to represent both the coarse-grained feature and the fine-grained feature. For building the connection between visual contexts and textual utterances, a light discriminative network is trained. The whole requirement for the discriminative network is to discriminate the degree of the connection between the given visual feature and textual feature. In each inference step, both the CV and FV model are required to generate N-best responses list with its probability as the forward probability rather than only the best response. And each response in N-best list along with the preceding visual feature are fed into the former trained discriminative network obtaining the backward probability. Finally, the forward probability and the backward probability is concatenated to rerank the N-best list. For more details please refer to \citet{wang2021modeling}.

\subsection{Results}
Following \citet{wang2021modeling}, we report the results in terms of the following automatic evaluation metrics: 
\begin{itemize}
    \item {\bf BLEU}: BLEU score is a common automatic evaluation method for majority NLP tasks \citep{papineni2002bleu, sordoni2015neural}, which score the n-gram overlaps between the generated sequences and reference sequences. For our experiment we report the BLEU-4 score.
    \item {\bf Diversity}: Diversity is usually reported in the task of dialogue generation \cite{li2015diversity}, which score the number of distinct n-grams in generated responses, and $n=1,2,3,4$ for this experiment. 
\end{itemize}
Results are shown in Table \ref{tab:bleu}. Since OpenViDial 2.0 is much larger than OpenViDial 1.0, we only use the top 5 objects for FineVisual model compared to using top 20 objects on OpenViDial 1.0, and this is the main reason why FV doesn't significantly perform better than FV and NV. 

\section{Conclusion}
In this report, we release OpenViDial 2.0, a larger-scale open-domain multi-modal dialogue dataset with visual contexts, updated from the previous version 1.0. 
OpenViDial 2.0 contains a total number of 5.6 million dialogue turns extracted from either movies or TV series from different resources,
and is four times larger than version 1.0 at scale. 
We hope this large-scale dataset can help facilitate future researches on open-domain multi-modal dialog generation.
OpenViDial 2.0 is available at \url{https://github.com/ShannonAI/OpenViDial}.

\bibliography{emnlp2020}
\bibliographystyle{acl_natbib}

\end{document}